\documentclass[
12pt,authoryear]{elsarticle}
\usepackage{hyperref}
\usepackage{graphicx,xcolor}
\usepackage{amssymb}

\journal{Artificial Intelligence}
\begin{document}
\begin{frontmatter}
\title{The Defeat of the Winograd Schema Challenge}
\author{Vid Kocijan\textsuperscript{a,1}}
\author{Ernest Davis\textsuperscript{b}}
\author{Thomas Lukasiewicz\textsuperscript{c,d}}
\author{Gary Marcus\textsuperscript{e}}
\author{\newline Leora Morgenstern\textsuperscript{f}}
\address{\textsuperscript{a}Kumo.ai, 357 Castro Street, Suite 200
Mountain View, CA 94041, United States}
\address{\textsuperscript{b}New York University, Department of Computer Science, 251 Mercer St, NY 10012, United States}
\address{\textsuperscript{c}Institute of Logic and Computation, TU Wien, Austria}
\address{\textsuperscript{d}Department of Computer Science, University of Oxford, UK}
\address{\textsuperscript{e}Robust AI, 380 Portage Avenue Palo Alto, CA 94306 United States}
\address{\textsuperscript{f}Palo Alto Research Center, 3333 Coyote Hill Rd, Palo Alto, CA 94304, United States}
\fntext[1]{Work performed while at the University of Oxford.}

\begin{abstract}
The Winograd Schema Challenge---a set of twin sentences involving pronoun
reference disambiguation that seem to require the use of commonsense
knowledge---was proposed by Hector Levesque in 2011. By  2019, 
a number of AI systems, based on large pre-trained transformer-based language models 
and fine-tuned on these kinds of problems, achieved better than 90\% accuracy.
In this paper, we review the history of the Winograd Schema Challenge and discuss the lasting contributions of the flurry of research that has taken place on the WSC in the last decade. We discuss the significance of various datasets developed for WSC, and the research community's deeper understanding of the role of surrogate tasks in assesssing the intelligence of an AI system. 
\end{abstract}
\begin{keyword}
Commonsense Reasoning \sep Winograd Schema Challenge
\end{keyword}
\end{frontmatter}
\section{Introduction}
\label{section-introduction}

In 2010, Hector Levesque (\citeyear{Levesque2011WinogradSchema}) proposed a new challenge for 
artificial intelligence: The Winograd Schema Challenge. The challenge
was named after a well-known example in Terry Winograd's~\citeyear{Winograd1972UnderstandingNL} 
ground-breaking doctoral thesis,  {\em Understanding Natural Language}. 
The example consists of a pair of sentences:

\begin{quote}
The city councilmen refused the demonstrators a permit because they
feared violence. \\
The city councilmen refused the demonstrators a permit because they
advocated violence.
\end{quote}

In the first sentence, the pronoun ``they'' 
is naturally interpreted as referring to the city councilmen; in the second,
it is naturally interpreted as referring to the demonstrators. The
only difference between the two sentences is that the first has ``feared'' where
the second has ``advocated''. Therefore, the different referents chosen for
``they'' must somehow reflect that different choice of word. Presumably when
humans read the first sentence, their choice of referent for ``they''
is guided by 
their knowledge that X fearing that Y would be violent would be a reason
for X to deny Y a permit to demonstrate, whereas Y fearing violence would 
rarely be a reason for X to deny Y such a permit. 
Humans' choice of referent in the second sentence is 
likewise guided by their knowledge
that Y advocating violence would be a reason to deny Y a permit to
demonstrate, whereas X advocating violence would not be a good reason for 
X to deny Y a permit to demonstrate. Knowledge about
the attitudes of city councils' vs.\ demonstrators' stereotypical attitudes
toward (non-state-sanctioned) violence no doubt also plays a role in
the disambiguation.  In 1972, and still in 2011, it seemed reasonable
to suppose that an AI that would be able to do this kind of disambiguation
would similarly have to draw on a body of commonsense knowledge.

\begin{table}
1972: Winograd's (1972) thesis introduces the original example. \\
2010: \cite{Levesque2011WinogradSchema} proposes the Winograd Schema Challenge. \\
2010--2011: The initial corpus of Winograd schemas is created. \\
2014: Levesque's  Research Excellence talk ``On our best behavior'' \\
\hspace*{2em}
\citep{Levesque2014BestBehaviour}. \\
2016: The Winograd Schema Challenge is run at IJCAI-16. No systems do \\
\hspace*{2em} much better than 
chance \citep{Davis2017WSC}.\\
2018: WNLI is incorporated in the GLUE set of benchmarks. BERT-based \\
\hspace*{2em} systems do no better than 
most-frequent-class guessing \\
\hspace*{2em} \citep{Wang2019Glue}. \\
2019, May: \cite{Kocijan2019MaskedWiki}  achieve 72.5\% accuracy on WSC273 using \\
\hspace*{2em} pretraining. \\
2019, June: \cite{Liu2019Roberta} achieve 89.0\% on WNLI. \\
2019, November: \cite{Sakaguchi2020WinoGrande} achieve 90.1\% on WSC273. \\
\caption{Time line of the Winograd Schema Challenge}
\label{tabTimeLine}
\end{table}

Levesque therefore proposed to use these kinds of sentences as a test to measure
the depth of understanding of AI natural language programs. In particular, 
he defined a {\em Winograd schema\/} as a pair of sentences,\footnote{In
later work, this was often relaxed, so that each element of the pair could
be a two-sentence text.}
comparable to
Winograd's example above, with the following features.

1. The two sentences are identical except for one or two words:
``feared'' vs.\ ``advocated'' in Winograd's example.

2. The two sentences both contain two noun phrases and a pronoun:
``the city council'', ``the demonstrators'', and ``they'' in Winograd's 
example.

3. The natural readings of the two sentences in isolation would assign
different choices of referents for the pronoun.

4. Simple feature matching, known as ``selectional restrictions'', will
not suffice to do the disambiguation. For instance, the pair of sentences
\begin{quote}
The women stopped taking the pills because they were [pregnant/carcinogenic].
\end{quote}
is disallowed, because pills cannot be pregnant and women cannot be 
carcinogenic.

5. Matching based on simple frequency of co-occurrence will not suffice
to do the disambiguation. For instance, the pair of sentences
\begin{quote}
The racecar zoomed by the school bus because it was going so [fast/slow].
\end{quote}
would be disallowed, because the words ``racecar'' and ``fast'' tend to 
appear together in text.

6. Both sentences must seem natural and must be easily understood by 
a human listener or reader; ideally, so much so that, coming across
the sentence in some context, the reader would not even notice the 
potential ambiguity.

Point (6) was taken for granted by~\citet{Levesque2011WinogradSchema} but was made explicit
in~\citep{Levesque2012WinogradSchema}. 

These features / conditions were assumed at the time to be straightforward to evaluate. It was also assumed that it would be relatively easy to write such Winograd schemas, so much so that it was originally suggested that a restricted vocabulary could suffice to generate large numbers of Winograd schemas, and that it would be possible to generate schemas that focused on particular areas of commonsense reasoning, which could enable focusing on measuring progress in these specific commonsense reasoning domains.

The first of these assumptions speaks to the suitability of Winograd schemas as a test for intelligence, substituting, as Levesque and later, his co-authors, argued, for the Turing Test; the second speaks to the feasibility of running the test. Both were crucial for the challenge's success. Indeed, one of the arguments for using the Winograd Schema Challenge rather than the Turing Test was the well-known difficulty of evaluating computer systems using Turing's criteria, a difficulty that has led to competitions of dubious worth, such as the Loebner competition, and the crowning of clearly non-intelligent agents like Eugene Goostman, as ``winners'' of the Turing Test \citep{levesque2017common}.  But neither assumption turned out to be correct. 

More precisely, regarding ease of evaluation: While it is easy to determine whether a computer system gets the correct answers on a test consisting of Winograd schemas (or more precisely, Winograd schema halves, as discussed below), it turns out that it is not easy to evaluate the quality of the test itself, that is,  whether these Winograd schemas satisfy conditions 1--6. Indeed, with the exception of conditions 1 and 2, these conditions can be difficult to evaluate, perhaps because they were not stated precisely enough,  For example (condition 4), it is often not clear whether a pronoun can be disambiguated using selectional restrictions, or what the boundary between this sort of disambiguation and commonsense knowledge is. It is also not simple to decide whether ``simple frequency of co-occurrence'' suffices to disambiguate. We need to specify: co-occurrence in which corpus or set of corpora? Do we use Google as the arbiter, as explicitly suggested in the paper, and if so how do we take into account the fact that Google can return different answers for different users depending on its history of interaction with them? And the fact that its corpora of reference are always changing?

Moreover, it is not necessarily easy to determine what the natural reading of a sentence is (conditions 3 and 6). The question of whether readers ordinarily would not  notice that the pronoun is ambiguous, has never been tested, as far as 
we know. (Reading-time experiments have been carried out for similar issues in the naturalness and coherence of texts; see, e.g.,  \citep{wolf2004discourse}.)  However, as discussed further in
Section~\ref{secConstructingWSs}, the constructors
of the dataset kept it very much in mind as a design consideration,
and in fact several examples that otherwise met the criteria were 
rejected, because the dataset's authors could not find any way to phrase them
that sounded natural. This was one of the factors that played into the difficulty of constructing a large set of Winograd schemas, in direct contradiction to our early assumption that such construction would be easy. 

Before the authors explicitly recognized just how difficult it would be to create sentences that satisfied conditions 3 and 6, the naturalness of the task was an aspect
of the Winograd Schema Challenge that was particularly striking as compared
to many other natural language tasks used as benchmarks. It made it
particularly compelling as a powerful test of human understanding. In a 
sentence from a well-designed schema, human readers could carry out the 
inference {\em automatically}, in a ``System~1''~\citep{kahneman2011thinking} process
in reading the sentence, {\em before\/} they read the question that points
out the ambiguity; although this inference seems to require commonsense reasoning
of some depth and complexity. 

This naturalness, if achievable, would result in a task that would be clearly superior to other tasks of comprehension and inference that had previously been suggested. For example, by contrast, characterizing the relation between two sentences as ``entailment'', ``neutral'', or ``contradiction'', as in the RTE task (recognizing textual entailment)  \citep{Dagan2006RTE}; or identifying the segment of a text that answers a question, as in SQuAD (Stanford Question Answering Dataset) \citep{rajpurkar2016squad}; or word analogies \citep{Mikolov2013word2vec}; or predicting a probability distribution over the next word following a specified text, which is the fundamental task in ``language modeling'' \citep{Brown2020GPT3}, are not ecologically valid tasks that people ordinarily carry out explicitly. 

As a contrast in a different direction, \cite{cozman2020winograd} propose encoding brainteasers as Winograd schemas; for example, 
\begin{quote}
Miss Marple was looking for the jewel, so she asked the girls about it:  Ann said she took the jewel, Bella said Donna was the thief, Carol said she did not even see the jewel, and Donna said Ann (did/didn't)
in fact take the jewel. When Miss Marple learned that only one of the girls was telling the truth, she immediately knew who had the jewel, and she smiled to {\em her}.
\end{quote}
It is not clear to us what is gained by casting this in terms of pronoun disambiguation, instead of simply asking the question; and it is certainly not what the creators of the challenge had in mind.

Levesque's original 2011 paper included eighteen original Winograd sche\-mas.
The first of these, ``The trophy doesn't fit in the brown suitcase
because it's too [small/large],'' has become a standard example, cited
hundreds of times in articles about AI language understanding.

Davis, Levesque, and Morgenstern intended that once a reasonable-sized collection of Winograd schemas had been assembled,
these could be used as a challenge for AI systems as follows. One sentence
of each pair would be chosen at random; that sentence would be turned
into a binary-choice question. For instance, the ``trophy'' schema would
yield the following question
\begin{quote}
The trophy doesn't fit in the brown suitcase because it is too large. \\
What is too large? \\
A. The trophy \\
B. The suitcase
\end{quote}

The advantage of using halves of Winograd schemas rather than simply using
sentences with pronoun ambiguity is that one can be sure that the only clue
to the correct resolution lies in the choice of the alternating word; any
clue that does not depend on the alternating word would point in the same direction
for both sentences, which would be the wrong direction for one of the pair.
For example, one can be entirely sure
that the ``recency'' heuristic has exactly a 50\% chance of working, since
it will work on one sentence of the pair, and fail on the other and the
sentence in the test set is chosen at random.

Thus, a program that guesses randomly will have an expected accuracy of 50\%.
Human readers should have accuracy close to 100\%. Thus, an AI program that
achieves an accuracy that is statistically larger than 50\% is accomplishing
something significant by way of understanding; an AI cannot claim to have
achieved human-level abilities unless its accuracy is close to 100\%.

Levesque argued that this test would be a better test of understanding
than the Turing test---trivial to evaluate, difficult to game. He also
argued that it would be more robust than some alternative semantic 
tasks that had been proposed, such as the RTE (recognizing textual 
echallenge~\citep{Dagan2006RTE}.

\section{Dataset creation}
\label{SectionDataset}

Building on Levesque's initial work, in 2010, Ernest Davis created an additional
eighty-nine schemas. Together with four suggested by Ray Jackendoff, these were
published on the web in HTML and 
XML.\footnote{https://cs.nyu.edu/faculty/davise/papers/WinogradSchemas/WSCollection.html and \\
https://cs.nyu.edu/faculty/davise/papers/WinogradSchemas/WSCollection.xml}
Since then, the collections have been amplified with new schemas, by Davis and
others; in particular, the collection includes twenty schemas written by David 
Bender. Currently, the HTML file
includes 150 schemas; the XML version includes 285 individual sentences. (Some
of the HTML schemas do not fit into the stricter form of the XML framework
and are thus not included.) The latter is generally 
known by the acronym WSC285 in recent articles on the subject; since it has
become a standard, it is now fixed and will not be modified further. 
To ensure consistency with earlier models, several authors often prefer
to report the performance on the first 273 examples only.  This dataset is known as WSC273.

Levesque, Davis, and Leora Morgenstern~\citeyear{Levesque2012WinogradSchema} published an expanded version of~\citep{Levesque2011WinogradSchema}, expanding the comparison to other proposed tests of 
AI semantic understanding and of the significance of the test in the
general context of AI research, as well as presenting the corpus. \citet{Levesque2014BestBehaviour} made the challenge the subject of his speech ``On Our Best Behavior''
accepting the 2013 IJCAI Research Excellence Award.

The spatial and physical knowledge used in disambiguating the trophy example---that proposed content $A$ will not fit inside container $B$ if either
$A$ is too large or $B$ is too small---is fairly complex from a logical
standpoint, particularly the interpretation of ``too''. (\citet{Davis2013SpatialReasoning} includes
a logic-based analysis.) Other Winograd schemas likewise draw on fairly
complex knowledge and reasoning. ``The sack of potatoes had been placed
[above/below] the bag of flour, so it had to be moved first" draws on
a different body of spatial and physical knowledge, and is similar to the knowledge needed to stack and unstack blocks in Blocks World. ``There is a pillar
between me and the stage, and I can't see [around it/it]'' draws on
knowledge of spatial relations and perception. ``Bob paid for
Charlie's college education, but now Charlie acts as though it never 
happened. He is very [hurt/ungrateful]'' draws on a body of knowledge
of human interactions and emotions, as well as the realities of the cost of college education. (Note that this example demonstrates that the quality of a Winograd schema must often be evaluated with respect to a specific culture. The example might not work so well in cultures in which college is nearly free.) ``Alice tried frantically to
stop her daughter from [chatting/barking] at the party, leaving us to wonder
why she was behaving so strangely'' requires reasoning about atypical
human behavior; for the variant with ``chatting'', this requires applying the
rather complex rule that ``It is generally inappropriate for
$A$ to try to prevent $B$ from carrying out an innocuous behavior $E$''. 
All told, the Winograd schemas draw on a wide range of knowledge, often
of fairly complex structure. Of course drawing on a wide range of knowledge does not entail knowing all, or most, or even a large part of the knowledge that one assumes an intelligent human would have. We believed that a system that knew and could reason with this knowledge would know even more and would have a means to acquire still more knowledge.

The challenge attracted a fair amount of favorable interest from both the
research community and the popular science press (e.g., \citep{markoff2015software,  knight2016tougher}). The naturalness of
the problems made it a fair one for AI systems; the complexity of the
inferences involved seemed to put it far beyond the range of what was then current 
technology. 

But even at the start and continuing throughout the early years of the challenge, when there were few signs that AI systems could succeed at the Winograd Schema Challenge, there were those who warned that it might
be a less robust test than its creators hoped and believed. They argued that, as was the case with past challenges, success might come without solving the underlying problem that had motivated the challenge, and moreover, that the very nature of the test invited systems that were tailored to the test rather than the underlying problem. In a
prescient email to Davis on Feb. 5, 2011, Doug Hofstadter wrote:
\begin{quote}
    The problem is that I believe that what will happen is that you
will simply wind up spawning a whole host of new and ultra-clever
brute-force techniques to solve the ``Winograd Challenge" without solving the
problem of understanding whatsoever.  I always liked Terry Winograd's sample
sentence, but I hardly think that it represents the epitome or the essence
of what is wrong with today's computer approaches to language.  It is just
one type of example among thousands of types.  It's a great example but it's
misleading to focus on it as if it were really the crux of the matter.
Getting people to spend huge amounts of time on just one kind of challenge
is not going to be helpful.  In fact, I fear it will be counterproductive,
because I don't think that anyone who will be moved to tackle this
particular challenge is likely to take up the deeper and more general
challenge of what language understanding really is.  People are daunted by
that, as well they should be, and no one is going to be motivated by a prize
to suddenly tackle that gigantic challenge.  Instead, very smart engineering
types are going to be motivated to seek clever tricks that will allow
computers to solve this very narrow type of linguistic disambiguation
problem with a high degree of accuracy. 
\end{quote}
A similar sentiment was expressed by Craig Boutillier at the KR-2012 conference after Morgenstern's presentation of the \citep{Levesque2012WinogradSchema} paper. He pointed out that challenges often do get solved sooner or later; and challenged the authors to a thought experiment. If a system did succeed at the Winograd Schema Challenge, would they be convinced of its intelligence through this alone? In a conversation with Morgenstern in April 2011 at a DARPA PI meeting for the Machine Reading program, Ed Hovy pointed out that this challenge was very likely the sort of problem that a smart graduate student could figure out how to solve, given the right resources. Indeed, the first author of this survey was until last year a graduate student who came up with the first system that performed much better than chance and achieved what was close to human performance, confirming Hovy's prediction. Dan Weld argued, at Marcus's AAAI-2015 workshop Beyond the Turing Test, that the nature of a narrow test was that researchers would figure out how to game the test without making real progress in the field. In addition, Charles Ortiz was one of several researchers who in 2013 suggested requiring systems to furnish an explanation for their choice of pronoun referent in order to guard against systems solving the WSC without having achieved the ability to do commonsense reasoning.

Hofstadter, Boutillier, Hovy, Weld, and Ortiz all turned out to be correct. Solving Winograd schemas is not a surrogate for the ability to do commonsense reasoning, let alone for intelligence. The difficulty of using success at a specific task as a surrogate for intelligence is one that continues to this day, even as the tasks that computers can successfully perform significantly increase in complexity. And the ability to furnish explanations has years later been recognized as a potential differentiator for determining the depth of understanding of systems that solve Winograd schemas

\subsection{Manually constructing Winograd Schemas}
\label{secConstructingWSs}
\citep{Levesque2012WinogradSchema} or any subsequent papers
did not discuss the actual process of constructing Winograd Schemas, because at that
time it seemed a reasonably easy task of no great interest. But the difficulty of creating more than a relatively small set of Winograd schemas (on the order of several hundred), the requests for larger sets,  the attempts to produce 
larger colllections of Winograd schemas using inexpert labor, and the proliferation of adversarial
data sets of all kinds in the last few years,  make it now 
worth a brief discussion.

The bulk of the Winograd Schemas in the online collection were constructed
by Davis over a month in the summer of 2010. Working an hour a day, he found
that he could reliably generate three schemas an hour --- rarely fewer, almost
never more. In the fall of 2015, preparing for the Winograd Schema competition
at AAAI-16, he produced another collection of 60, which has been held in 
reserve and has never been published. This was carried out less systematically,
and was somewhat slower, but not enormously so. Overall, the continued slow pace led us to the development of the
Pronoun Disambiguation corpus, discussed~below.

The primary difficulty in constructing Winograd schemas is that English usage
imposes a variety of formal constraints on how pronouns can be used, beyond purely syntactic rules. These constraints are complex and not fully understood \citep{Rohde2018PronounInterpretation},
but can be quite rigid, nonetheless.

For instance, both halves of Levesque's canonical trophy example are fine:
\begin{quote}
The trophy doesn't fit inside the suitcase because it is too large. \\
The trophy doesn't fit inside the suitcase because it is too small. 
\end{quote}
Commonsense knowledge about fitting and size require that, in the first sentence,
``it'' refers to the trophy, and in the second ``it'' refers to the suitcase.

But you can hardly say
\begin{quote}
*   The trophy doesn't fit in the suitcase because the trophy is an awkward shape and it is too small. 
\end{quote}
with the intent that ``it'' should refer to the suitcase. 

Likewise, you can reasonably say in English,
\begin{quote}
Ann has no children, but Barbara has two sons, Carl and David.
Her children are in elementary school.
\end{quote}

However, you cannot say
\begin{quote}
* Barbara has two sons, Carl and David, but Ann has no children.
Her children are in elementary school. 
\end{quote}
in the expectation that semantic constraints will force the interpretation of ``Her'' as ``Barbara's''. The natural reading of this sentence is to interpret ``Her'' as meaning ``Ann's'' and then try to work out the contradiction.

\cite{kehler2015testing} similarly gives a number of examples where the form of the sentence either forces a pragmatically implausible reading or conflicts with pragmatics in a way that makes the sentence difficult to understand:

\begin{quote}
 (a) The demonstrators were denied a permit by the city council because they
feared violence. \\
(b) The city council denied the demonstrators a permit because they felt
strongly that the best way to draw attention to current political issues is to
advocate violence.\\
(c) Norm lent his car to his brother’s girlfriend. He doesn’t own one. \\
(d) Margaret Thatcher admires Hillary Clinton, and George W. Bush absolutely
worships her.
\end{quote}

In (a), readers generally interpret ``they" as referring to the demonstrators. In (b), readers ``garden path": they initially interpret ``they" as meaning ``the city council" and then consciously backtrack and correct when they reach the end of the sentence. In (c), readers end up confused rather than deciding that ``he" refers to the brother. In (d), readers interpret ``her" as meaning Clinton rather than Thatcher, despite the fact that, when the example was devised,
Thatcher was much more plausible.

Since the two sentences in a Winograd schema can differ only in a word or two,
and both sentences have to be very easily understood by a human reader, and
both have to sound natural in their use of pronouns, and 
neither can be resolvable by selectional restrictions or considerations
of frequency,~\footnote{It has become very clear that the measure of frequency originally considered for the WSC, namely, the number of results returned by Googling phrases, without any consideration of context, is much too simple a notion.}  
the problem of generating Winograd schemas is substantially overconstrained.
The problem of generating a collection of Winograd schemas that span a range
of commonsense knowledge and linguistic constructions is still harder. Often
it happened that a promising first sentence could not be completed to a 
full schema. For instance, the text ``Tom's books are full of mistakes.
Some of them are quite [foolish/worthless].'' seemed promising; but
``foolish mistake'' is a much more common association than ``foolish book'',
and we couldn't find a way to fix this.  

Another example: The following sentence is from a published article \citep{Joukovsky2011Peacock}:   
\begin{quote}
Late in 2009, the novelist Jim Powell found a cache of letter written 
by the 19th century novelist George Meredith to his great-great-grandmother 
Susan Mary Neil.  
\end{quote}
The pronoun ``his'' here most likely refers to Powell, not Meredith; very few people in the nineteenth century
were in a position to write letters to their own great-great-grandmother.

Even setting these issues aside, the Powell example could be turned into a Winograd schema by changing 
``great-great-grand\-mother'' to ``friend''; in that version, ``his'' must
refer to Meredith, since it would be unusual (though not quite impossible)
for a nineteenth-century novelist to write letters to a friend of someone alive
in 2009. However, the problem with this example is that the first version is 
too easily misread. In fact, Davis, reading the article, {\em did\/} 
misinterpret it until he got further in the article and realized that
interpreting ``his'' as ``Meredith's'' was not only wrong but close to impossible.

There is a web 
page\footnote{https://cs.nyu.edu/faculty/davise/papers/WinogradSchemas/WSfailed.html}
with a few more examples of attempts at Winograd schemas that could not
be brought to fruition.

Manually creating a large, diverse collection of high-quality Winograd schemas is inherently difficult. As a result, the collection was small; initially a little more than a hundred, and later less than three hundred. This led to some complaints; for instance,  \cite{Trichelair2018WSCAnalysis} wrote that ``the main drawback of the Winograd Schema Challenge is its limited size and the absence of training and validation sets for (hyper)parameter tuning. For its size, it turns out that if one were to choose from a set of 10 random binary classifiers, the best based on its performance on the WSC, there is more than a 1-in-3 chance of scoring above 55\% accuracy with this chosen classifier.'' 
However, it was not our purpose to construct a training set, because there is no point in training an AI program to solve Winograd schemas specifically. The point of the Winograd Schema Challenge is to test programs that claim to have solved the problem of pronoun reference resolution; and for that purpose, what would be relevant would be a training set for pronoun reference resolution generally. As for testing for pronoun resolution: a sample of 100 would be small for polling voters in an election, where you generally expect percentages between 40\% and 60\% and you would like to get precision of around 2\% or 3\%. But for distinguishing between success rates for humans on the Winograd schemas, which are around 95\% and those of 2011-level AIs, which were little better than 50\%, a sample of 100 is more than adequate. If an AI scores 50\% on a sample of 100, then one can reject the hypothesis that its actual accuracy is 95\% with a confidence of around $1-10^{-36}$. (\citet{linzen2020can} similarly argues for the value of expert-constructed, comparatively small, ``test-only'' datasets.)

However, if we look beyond the problem of pronoun resolution and consider the fact that this problem was intended to be a surrogate for commonsense reasoning, we note that it is still the case that a set of 100 or so Winograd schemas is unlikely to cover or even touch the many domains and subdomains and aspects of commonsense reasoning that would be expected to be in the purview of an intelligent agent.

\subsection{Subclasses of the original collection}

\citet{Trichelair2018WSCAnalysis} have observed that $37$ sentences in the \textsc{Wsc273} dataset ($13.6\%$) can be easily solved using statistics over simple patterns; that is, they fail condition (5) above.
An example of such a sentence is 

\textit{In the storm, the tree fell down and crashed through the roof of my house. Now, I have to get \textbf{it} [repaired/removed].}

\textit{The roof} is commonly associated with \textit{being repaired}, while \textit{the tree} is not.
They call these examples \textit{associative} and name the rest \textit{non-associative}.

Such patterns can involve multi-word phrases. For instance, the example

\textit{Joe has sold his house and bought a new one a few miles away. He will be moving out of \textbf{it} on Thursday.} 

\noindent
can be solved using the fact that \textit{moving out of his old house} is a much more common phrase than 
\textit{moving out of his new house}. (As of May 13, 2021, a Google search claimed that there were 481,000 results for the first quoted phrase and 2 for the second.)
\citet{Trichelair2018WSCAnalysis} used human annotators to identify the associative sentences; their identifications were largely replicated by \cite{elazar2021back}, using an automated measure.

%
%
%

\subsection{Pronoun Disambiguation Problem Dataset}

The Pronoun Disambiguation Problem (PDP) dataset consists of 122 problems of pronoun disambiguation collected from classic and popular literature, newspapers, and magazines. As discussed by~\citet{morgenstern2016planningwsc}, PDPs were originally conceived as an auxiliary dataset to Winograd schemas primarily because constructing Winograd schemas according to Levesque’s original guidelines was a difficult, manual process. Indeed, as of this date, the plurality of such Winograd schemas has been constructed by Davis. Moreover, in the decade since the Winograd Schema Challenge was first published, no one has successfully developed a methodology for constructing large numbers of such schemas. There have been efforts to construct large sets of Winograd schemas, such as \textsc{WinoGrande}~\citep{Sakaguchi2020WinoGrande}, but as discussed in \ref{Section:Winogrande}, the schemas are often flawed and do not come close to meeting Levesque’s guidelines. 

The chief distinction between PDPs and Winograd schemas is the lifting of Levesque’s criterion of the ``special word’’ or phrase that when substituted creates a pair of similar sentences with different pronoun referents. Winograd schemas are rarely found ``in nature,’’ that is, in novels, short stories, newspaper articles, or other pieces of existing text. In contrast, PDPs are ubiquitous in text.
Not every instance of a pronoun in text, of course, is a suitable PDP. For example, the sentence starting Chapter 4 in Maud Hart Lovelace’s {\em Carney’s House Party} ``During the long trip Carney began to think about her reunion with Bonnie,” has the pronoun {\em her}, but there is no challenge in resolving that pronoun to {\em Carney}. A suitable PDP should ideally present a slight challenge to a human reader, a challenge that is easily resolvable using commonsense knowledge, but a challenge nonetheless. An example, from Alcott’s {\em Little Women} is:
\begin{quote}
Mrs.\ March gave the mother tea and gruel, while she dressed the little baby as tenderly as if it had been her own
\end{quote}

There are two possible referents for \emph{she}: \emph{Mrs.\ March} and \emph{the mother}. It is not difficult  to reason that \emph{she} must refer to Mrs.\ March rather than the mother, since it is clear from the remainder of the sentence, especially the phrase ``as if it had been her own’’ that the little baby does not in fact belong to the referent of \emph{she}. Humans are likely to do this reasoning using, among other things, the commonsense knowledge that a baby is understood to belong to its mother.

The aim in collecting and vetting PDPs has been to come up with a set consisting of challenging problems that tests the ability to do something akin to commonsense reasoning. They were intended to be used as a gateway set before administration of the Winograd Schema Challenge and in fact were used in this way in the IJCAI-2016 running of the Challenge~\citep{Davis2017WSC}. Our aim in construction and vetting was to create a test that was roughly equivalent in difficulty to a WSC test corpus, though of course, the absence of pairs of sentences distinguished only by a single word or phrase means that one cannot guarantee that the pronoun reference is resolved through commonsense knowledge and reasoning ability, rather than sentence structure.

PDPs are clearly easier to collect and to vet than Winograd schemas are to construct. In addition, there are other reasons that appropriate sets of PDPs are of interest. First, by construction, Winograd schemas are generally quite simple --– in general, the more complex the sentence, the more difficult it is to create a sentence identical in all but a word or short phrase ---  and it may be of interest to test a system’s ability to resolve pronoun referents in contexts that are more complex. 

Second, the reasoning and the knowledge involved in many PDPs has tended to be deeper and involve more types of commonsense knowledge than that needed for Winograd schemas.  An example of a PDP of this sort is taken from \emph{All-of-a-Kind Family}:
\begin{quote}
Mama came over and sat down beside Sarah. Gently she stroked her hair and let the child weep.\end{quote} 

This example involves relatively profound concepts of parenting and empathy and comforting and soothing that are not, at least at this point, present in existing Winograd schema collections, perhaps partly because it is not straightforward to find a sentence that is nearly identical to the one above, but for which the referent for \emph{she} would be Sarah rather than Mama. Such a range of difficulty and complexity of reasoning in Winograd schemas was a desideratum in Levesque’s original paper.
Finally, finding pronoun referents in ``naturally occurring’’ PDPs corresponds to a useful task in a way that finding pronoun referents in Winograd schemas do not, precisely because the latter are artificial.

Despite the fact that collecting and vetting PDPs is a much simpler task than constructing Winograd schemas, it is not a trivial task. Morgenstern et al.\ focused on children’s books partly because they tended to contain a wide range of commonsense reasoning concepts of interest. However, some candidate examples did not stand up to careful examination; others were trivial or did not involve interesting or complex commonsense concepts.

62 examples of PDPs were published before the Winograd Schema Challenge was administered,\footnote{\url{http://commonsensereasoning.org/disambiguation.html}} and 60 PDPs were included in the Winograd Schema Challenge that was administered at IJCAI 2016  \citep{Davis2017WSC}.\footnote{\url{https://cs.nyu.edu/faculty/davise/papers/PDPChallenge.xml}}
A~corpus of 400 sentences was collected semi-automatically from online text, with less vetting by Davis and Pan~(\citeyear{Davis2015PDPlarge}).\footnote{https://cs.nyu.edu/faculty/davise/annotate/corpus.xml} In addition, in developing this set, there was little emphasis on collecting examples that, for pronoun resolution by humans, would involve complex concepts of commonsense reasoning.

\subsection{Other collections of Winograd Schemas}
\label{SectionOtherCollections}
Over time, additional collections of Winograd Schemas or closely related examples have been created, either for the purpose of creating additional examples (e.g., for training), or to use them for purposes other than commonsense evaluation.
In this section, we discuss different approaches to construction of more schemas and their uses, with a detailed discussion of all existing datasets provided in \ref{other-datasets-appendix}.
These datasets can be grouped into five types, outlined below. The first, third, and fourth types are of particular interest in their potential to have a lasting effect on various research communities, independent of whether or not they have helped progress systems that can correctly solve Winograd Schemas.

The first category includes attempts at creating larger collections of examples, often by using inexpert labour and automatic filtering.
Examples of such construction procedures are, for example, asking students for help~\citep{Rahman2012DPR}, extracting examples from literature and manually vetting them~\citep{morgenstern2016planningwsc, isaak2020winventor}, or completely relying on inexpert labor from crowdsourcing platforms~\citep{Sakaguchi2020WinoGrande, Isaak2019WinoFlexi}.
Depending on the amount of expert labour involved, these datasets can be constructed at a much larger scale, usually at the cost of example quality.
Examples in the datasets of this type often fail to meet one or more of the features listed in Section~\ref{section-introduction}. Nevertheless, some work in this area, including the crowdsourcing approach taken in WinoGrande is of interest, because it is part of a growing movement to massively crowdsource examples when large datasets are needed. For example, the crowdsourcing methods used in WinoGrande have also been used in developing other datasets developed by Choi's group at AI2, such as Social-Chem-101. Understanding the ways in which crowdsourcing fell short in WinoGrande has helped us understand anomalous entries in Social-Chem-101.

Second, there are translations into other languages. These often come with various linguistic challenges as Winograd Schemas translate into co-references other than pronoun resolution~\citep{Bernard2020Mandarinograd,zagar2020WSCslovene}.
Moreover, in languages with a strong presence of grammatical gender,  modifications are necessary to avoid giving away the answer~\citep{Melo2020WSCportugese, Amsili2017WSCfrench}.

However, not all Winograd Schema collections were created with an intent to measure commonsense reasoning capabilities.

The third category includes datasets that aim to measure the impact of gender bias on model predictions. 
Winograd Schemas that describe gender-stereotypical (female homemakers, male CEOs) or anti-stereotypical (female astronauts; male domestic workers) scenarios can be used to observe how the gender of the pronoun affects the prediction.
Datasets of this kind have been constructed to detect gender bias in systems for coreference resolution~\citep{Rudinger2018WinoGender, Zhao2018WinoBias} and machine translation \citep{stanovsky2019BiasMT, emelin2021winox}. This work remains relevant even in an age of large language models (LLMs), since much of data used to train these LLMs is steeped in gender bias. While obviously an LLM that has been trained on a specific small set of Winograd schemas would answer those questions correctly, this fine tuning could itself be useful in reducing the gender bias of an LLM. 

Fourth, there are datasets that additionally come with explanations. The aim of these datasets is not only to evaluate model correctness, but also to evaluate its ability to provide a correct explanation, usually in natural language~\citep{zhang2020winowhy,he2021winologic,Yordan2021eWinogrande}. AI researchers had suggested requiring systems to furnish explanations of why pronoun referents were chosen since shortly after the WSC was published, as discussed earlier. 
But datasets supporting explanations became particularly relevant after the human performance on the Winograd Schema Challenge was reached.
Models that can make the correct prediction are often unable to provide or even pick the correct explanation for their prediction.


Fifth, and most distant from the original, Winoground \citep{thrush2022winoground} is a challenge dataset for visio-linguistic compositional reasoning. Each example consists of a pair of images and a pair of captions that differ only in word order; the task is to match the correct caption with the correct image (e.g., \emph{ a lightbulb surrounding plants} as opposed to \emph{plants surrounding a lightbulb}). 

\citep{storks2019recent} is an extensive survey and analysis of benchmarks for natural language inference, through 2019. 


\section{Methods}
\label{SectionMethods}
At least three different methods have been used to try to solve the Winograd Schema Challenge. One class of approaches consists of feature-based approaches, typically extracting information such as semantic relations~\citep{Rahman2012DPR,Peng2015WinoCoref}.
Additional \textit{commonsense knowledge} is usually included in the form of explicitly written rules from knowledge bases, web searches, or word co-occurrences~\citep{Emami2018KnowledgeHunter}.
The collected information is then used to make a decision, using rule-based systems, various types of logics, or discrete optimization algorithms~\citep{Isaak2016WinoSense}.
We observe that the extraction of relevant information from the sentence is usually the bottleneck of these approaches.
Given the nature of the challenge, even the slightest noise in the feature collection can make the problem unsolvable.


The second group of approaches are neural approaches, excluding language-model-based approaches, which we consider as a separate group.
Neural-network-based approaches usually read the sentence as a whole, removing the bottleneck of information extraction~\citep{Opitz2018WSCRanking}.
To incorporate background information, these networks or their components are usually pre-trained on unstructured data, usually unstructured text, or other datasets for coreference resolution~\citep{Liu2017KEE}.
Common approaches to the tasks in this group take advantage of semantic similarities between word embeddings or use recurrent neural networks to encode the local context~\citep{Zhang2018DistributedWSC,Wang2019UDSSM}.
We find this group of approaches to lack reasoning capabilities, as semantic similarity or local context usually do not contain sufficient information to solve Winograd schemas.

The third group includes approaches that make use of large-scale pre-trained language models, trained with deep neural networks, extensively pre-trained on large corpora of text.
Some of the approaches then additionally fine-tune the model on Winograd-Schema-Challenge-style data to maximize their performance~\citep{Kocijan2019MaskedWiki, Ruan2019WSCNSP}.
Approaches in this group achieve visibly better performance than approaches from the first two groups and led to the eventual near-human performance on all benchmarks~\citep{Lourie2021Unicorn, Sakaguchi2021WinoGrande}.

All approaches are compared and discussed in \ref{appendix-methods}.
The ability of AI systems to achieve a high accuracy on Winograd Schema benchmarks went from random guessing to human performance in a very short time span, a phenomenon described as an ``emergent capability''~\citep{Wei2022emergent}.


\subsection{The Winograd Schema Challenge at IJCAI-16}
The first and last running of the Winograd Schema Challenge as a competition took place at IJCAI-16 \citep{Davis2017WSC}. Six systems were entered and tested on a collection of 60 PDPs. The most successful of these was that of \cite{Liu2017KEE} discussed in \ref{section-Neural}, which achieved a score of 58\% on the test collection. The other contestants for the most part used knowledge-based technology, and submitted systems that incorporated only the specific knowledge that was required for the examples that had been published. Not surprisingly, these systems did not do better than chance when tested on new examples.

\section{The Problem of Commonsense Reasoning Remains}
\label{secCommonsense}
The Winograd Schema Challenge as originally formulated has largely been overcome. However, this accomplishment may in part reflect flaws in its formulation and execution. Indeed, \cite{elazar2021back} argue that the success of existing models at solving WSC may be largely artifactual. They write

\begin{quote}We provide three explanations for the perceived progress on the WS task: (1) lax evaluation criteria, (2) artifacts in the datasets that remain despite efforts to remove them, and (3) knowledge and reasoning leakage from large training data. 
\end{quote}

In their experiments, they determined that, when the form of the task, the training regime, the training set, and the evaluation measure were modified to correct for these, the performance of existing language models dropped significantly. 

The problem of commonsense reasoning still stands as one of the major challenges facing AI \citep{davis2015commonsense}. Large language models often give answers that are nonsensical or just plain wrong~\citep{Marcus2020GPT3}, as do ques\-tion-answering systems. Speech transcription and auto-correct systems of every kind regularly turn sensible speech and text into gibberish, by turns confusing, humorous, or embarrassing. Vision systems do not ensure that their interpretations are physically or even geometrically coherent. Video interpretation systems often have little idea of what is happening in a video. Robotic systems are extremely limited in their understanding of the world that they are interacting with. While no computer system or human's performance is perfect, the frequency and magnitude of system error is too great for such systems to be considered reliable without a great deal of help. This is a serious concern for technologies, such as self-driving cars, in which near-perfect performance is crucial in order for the technology to be considered useful, reliable, and ethical.  

This gap has recently come increasingly to the fore of the general discussion of AI challenges. For example, over the last decade, DARPA has created several large programs in different aspects of commonsense reasoning, including Causal Exploration, Computational Cultural Understanding, and Machine Common Sense \citep{gunning2018machine}. Commonsense has become a hot buzzword, as can be seen by its multiple mentions in the popular press (e.g., New Uorker, Hutston, April 5, 2022). Numerous commonsense benchmarks have been created; a collection maintained by Davis\footnote{\url{http://cs.nyu.edu/~davise/Benchmarks/}}, as of the time of writing, contains 157 such benchmarks, and new ones are published frequently.  However, it still remains the case that research into commonsense reasoning constitutes a small fraction of AI research---indeed, a rapidly shrinking fraction, given the extraordinary explosion of AI research---and that, almost without exception, successful AI applications avoid the issue.

Meanwhile, the need for commonsense understanding in AI systems remains as crucial as ever. The facts that, if a small object $O$ does not fit into a container $C$, then $O$ will not fit into a container smaller than $C$, and an object larger than $O$ will not fit into $C$\footnote{Depending on the exact interpretation of ``larger'' and ``fit'', this can be either a reliable inference or a plausible inference.} are the same whether the object and container are a trophy in a suitcase, a picture on a printed page, or a quart of milk in a bottle. These facts are the same whether they are being used to disambiguate the ``trophy'' sentence; to understand a narrative or video in which a container is being packed; to plan how to load a container; or to write a program for a robot to load containers. They can be used metaphorically in reasoning about fitting data into computer memory or fitting content into an article with word limits. They are part of a rich, general understanding of space and motion and of the constraints on how various kinds of physical objects can be moved and manipulated \citep{davis2017commonsense, lake2020word}. The fact that large language models with one hundred billion parameters trained on half a trillion words can learn enough linguistic patterns that they can disambiguate the pronouns in the ``trophy'' sentence does not solve the larger problem reliably; it is not even guaranteed to be progress toward reliably solving the larger problem.

 The WSC and similar datasets do not in fact explore all aspects of the use of commonsense knowledge in resolving pronoun references, although originally it was hoped that the WSC would do precisely that. For one thing, these datasets are limited to referents explicitly named in the same or previous sentence. In ordinary usage, the referent of a pronoun may be implicit from the context but not previously named in any noun phrase; for example, the referent of {\em it} in {\it I hiked for an hour and it tired me out} or the referent of {\it they} in {\it I went to the hospital but they sent me home.}
 
Furthermore, as argued in \citep{elazar2021back}, the success at a particular task, such as pronoun disambiguation of a model that has been fine-tuned to that task is not at all a reliable measure of the degree to which the model has learned commmonsense knowledge broadly, or even to which it has learned the commonsense knowledge needed for language understanding. For one thing, pronoun disambiguation may only probe a limited subset of the commonsense knowledge needed in natural language understanding. For another, the fine-tuned model may only able to use its commonsense knowledge for this single task but not for other understanding tasks. For instance, it may ``know'' that small objects fit better in large containers than vice versa for the purpose of pronoun reference resolution, but not for any other kind of disambiguation or interpretation.
We do note that \citep{elazar2021back} focused
on systems trained with supervised learning; their analysis does not
touch on the more recent successes of systems trained with unsupervised
learning only. 

The goal of the WSC was to present a challenge that (a) was clearly ``commonsensical'' and easily carried out by humans; (b) was easy to evaluate; (c) was an adequate test of commonsense reasoning abilities. It succeeded at (a) and (b), but not at (c). 

We are  doubtful that any benchmark or challenge set that consists of many bite-sized instances of a single simple tasks can adequately test for commonsense reasoning. As seen in Section~\ref{SectionOtherCollections}, it may always be possible to construct a training set comparable to the dataset and to train a system that can exploit task-specific features, while still avoiding confronting the commonsense knowledge that is involved when humans do the task---especially if, as is general practice, the training set and the test set are random samples of the same corpus.
As seen in Section~\ref{SectionMethods}, combining such datasets with models that can fit increasingly larger corpora, and/or augmenting them with task-specific tricks, it seems to always be possible to obtain a model that achieves high accuracy on examples of such a narrowly-defined task.

Rather, it may be necessary to relax one of the criteria originally considered to be essential to the WSC: that of a clear cut evaluation measure. It may be necessary, for example, to look at larger tasks with less tightly-structured measures of success, such as reading an extended document and giving free-form answers to free-form questions, or carrying out a new task from verbal instructions.\footnote{There are, currently, almost no commonsense benchmarks
involving free-form Q/A. Among the 48 tasks tagged as ``common sense'' in the Big-Bench collection, only one, ``{\tt tellmewhy}'' allows free-form answers to a limited class of questions \url{https://github.com/google/BIG-bench/blob/main/bigbench/benchmark_tasks/tellmewhy}. Among the other 56 text-oriented 
benchmarks in the collection maintained by Davis, none are Q/A system that allow free-form answers. \url{https://cs.nyu.edu/~davise/Benchmarks/Text.html}}. Of course, allowing free-form answers makes it difficult to accurately carry out automated evaluation.
An alternative approach would be to interactively probe the AI system's depth of understanding. In designing such a measure, it is always important to keep in mind that, at least at present, AI tends to stumble over basic concrete realities much more than over abstractions; it is more likely to fail conspicuously in discussing a pot being knocked off a counter than in discussing whether automata can be truly conscious \citep{davis2021using}.

\section{Larger Lessons}

AI challenge problems other than the Winograd Schema Challenge have been raised, studied, and solved without substantially changing the general AI landscape. Examples of such challenge problems include checkers (mastered by Chinook), chess (mastered by Deep Blue), and go (mastered by AlphaGo). All introduced new techniques; some like AlphaGo used novel combinations of self play and reinforcement learning techniques that have greatly influenced the AI research community – and have made possible, for example, the recent progress in automated playing of Diplomacy \cite{meta2022human}.  But although they solved difficult problems, these were not problems that were believed to be essential to the development of an artificially intelligent agent. After all, most humans do not excel at checkers, chess, or go, but are no less intelligent because of this lack.

The Winograd Schema Challenge was different in that it targeted pronoun reference disambiguation, a task that seems simple for humans to perform and is widely performed by people in  nearly all cultures since the development of spoken language. The actual target was larger still: the ability to do commonsense reasoning, an ability that nearly all humans have to some extent. At the same time, it maintained the definitive evaluation criteria that distinguish the best AI challenge problems: it is easy to tell when they have been solved.

Because the expectations were higher – if this challenge were solved, we believed, we would be able to develop artificial agents that were capable of tasks that have been considered unique to human intelligence – its defeat is sobering. 

What does it mean when a carefully constructed challenge that seemed to combine both an important target and rigid evaluation criteria is solved without in fact reaching the larger target?

We believe that there are four general lessons to be learned:

 \noindent
 \paragraph{\textbf {1. We need to be cautious when introducing proxy problems}} The use of proxy problems has been common to AI since its inception: The Turing Test, which is precisely concerned with developing a proxy test for an artificially intelligent agent, was formulated just as the field of artificial intelligence (not yet labeled as such) was being founded. To some extent nearly all tests are proxies. When one gives a test to make sure that an individual knows how to solve a problem or do a task, the test necessarily only validates a part of  that problem or task. Proxies are not troublesome if the gap between the test and the targeted problem is reasonably small.

But if there is a large gap, as there is between the Turing Test and the task of acting as intelligent as a human, or between the Winograd Schema Challenge and the task of performing commonsense reasoning, that proxy problem is an invitation to disappointment. The statement of a proxy problem  rarely includes a solid argument establishing the connection between the two problems. In the case of the Winograd Schema Challenge, there were in fact two proxy problem statements: first, that if a system solved the Winograd Schema Challenge, it would have a reasonable amount of commonsense knowledge and be able to perform commonsense reasoning, and second, that if a system solved the Winograd Schema Challenge, it could also act in an intelligent manner, akin to what is described in Turing's original formulation of the Turing Test.

Although good arguments were made to bolster these proxy problem statements, they were not solid enough, and both turned out to be false. 

This does not mean that solving challenge problems or excelling at a well-constructed test are meaningless achievements. But we need to be much more careful about what precisely they do mean. Note that this applies also to recent AI developments like DALL-E and ChatGPT. Their abilities, even if consistent, are not proof of the existence of artificially intelligent agents.

\noindent
\paragraph{\textbf{ 2. We need to be careful not to rely on a perceived connection between tasks and methods}} We should not fall into the pitfall of assuming that if humans accomplish task X using method Y, then non-human entities, including computer systems, must also accomplish task X using method Y.\footnote{Note that this is a distinct problem from assuming that Problem Z is a proxy for Problem X. Here were are concerned with assuming that a specific \emph{method} for a particular task is the only method that works. }

It was very tempting to believe, as we developed Winograd schemas or selected Pronoun Disambiguation Problems, that being able to determine a pronoun’s referent depended on having the requisite commonsense knowledge and being able to use that knowledge. We have given multiple examples in this paper of the sort of knowledge and  reasoning that we envisioned, e.g., understanding that funding a person’s college education is a generous action, and that the receiver is ordinarily quite grateful for the largesse.

But we ignored a fundamental truth: Different human beings solve problems in different ways, and certainly, computer systems may solve problems in different ways than humans do. What seems an obvious, even essential element of a solution to some set of humans may be neither. In the case of the Winograd Schema Challenge, it was a mistake to assume that commonsense knowledge and the ability to reason with that knowledge are prerequisites for pronoun disambiguation. In contrast to our expectations and predictions, machine learning methods combined with very large datasets and large language models do very well at pronoun disambiguation. Nevertheless, there is little evidence to support any conclusion they have the ability to do commonsense reasoning consistently well.

\noindent
\paragraph{\textbf{3. We should be careful to create test sets that span, as much as possible, instances of the targeted problem}}
As discussed earlier, due to the intrinsic difficulty of constructing Winograd schemas, it was very difficult to start out with a target domain of commonsense knowledge and to construct Winograd schemas that corresponded to that domain. The existing set of Winograd schemas touch on physical, spatial, interpersonal, and social reasoning. But except for one Winograd schema, they do not address nested modalities, e.g, the ways in which individuals reason about their knowledge of others' beliefs or lack of knowledge. They do not touch on ethical issues, such as whether sacrificing one life to save many others is appropriate, or how to allocate scarce resources and to balance needs against wants. They hardly touch on the commonsense knowledge specific to crafts like needlework and knitting (including, for example, the way rigid objects and flexible objects interact, or that specific arrangements of flexible objects can become denser and less flexible) or to agricultural activities or to many other domains. 

Narrow test sets are often useful, and it is often necessary to start with a smaller than desired test set in order to make progress in research. Moreover, it is possible and perhaps even likely that systems that solve
Winograd schemas using large language models would also successfully solve schemas in domains not present in the collection. However, the fact that these other schemas are not present means that even if we were to ``fix'' the WSC, solving the challenge would at best show proficiency for the problems in the domains covered by the test sets. It would not suffice to demonstrate proficiency in general commonsense reasoning.

\paragraph{\textbf{4. We should recognize that there is often a tension between tests that are easy to evaluate and tests that are definitive}}   One of the most appealing features of the WSC was the ease of evaluating whether or not a system passed the test. But, as is often the case with easy-to-evaluate multiple choice exams, even for humans, one can learn to do well on the test without being particularly proficient in the tested domain. We decided not to include as a requirement for solving the WSC the ability to furnish explanations, because  there was no clear way to evaluate such explanations. As a result, good performance on the Winograd Schema Challenge on its own ended up being a weak and unconvincing proof of commonsense reasoning ability, if at all, proving the concerns that Hofstadter and Boutillier expressed in 2011 and 2012 (see Section~\ref{SectionDataset}) to be correct. The decision to not require any explanation or justification with the prediction may have hastened the defeat of the Winograd Schema Challenge, though if we had decided in favor of requiring explanations, we would probably now be writing about the 
difficulties of performing the evaluation task and the significance of the results.

We believe that these larger lessons, if kept in mind by AI practitioners when they pose challenge problems in the future or try to determine the significance of new AI technology, will help clarify the state of current AI research and pave the way for meaningful AI research in the future.

\appendix
\section{Overview of Other Collections of Winograd Schemas}
\label{other-datasets-appendix}
\subsection{Winograd Schema Challenge in other languages}

While the inspiration and original design of the challenge was in English, translations into other languages exist. 
Amsili and Seminck~(\citeyear{Amsili2017WSCfrench}) translated the collection of $144$ Winograd schemas into French, and $285$ original Winograd schemas were translated into Portuguese by~\citet{Melo2020WSCportugese}.
Authors of French and Portuguese translation both report having to make some changes to the content to avoid unintended cues, such as grammatical gender.
In the case of Portuguese, $8$ sentences had to be dropped, as no appropriate translation could be found.

Translations to Japanese\footnote{\url{http://arakilab.media.eng.hokudai.ac.jp/~kabura/collection_katakana.html}} and Chinese\footnote{\url{https://cs.nyu.edu/faculty/davise/papers/WinogradSchemas/WSChinese.html}} are available on the official web page of the challenge. A second translation into Chinese, 
``Mandarinograd'',\footnote{\url{https://gitlab.com/vanTot/mandarinograd/}} is reported in \citep{Bernard2020Mandarinograd} together with an account of the difficulties involved in the translation process. 

 The Chinese Winograd Schema Collection (CLUEWSC2020) 
is an ana\-pho\-ra/coreference resolution task where the model is asked to decide whether a pronoun and a noun (phrase) in a sentence co-refer (binary classification). The collection contains 1838 questions hand-selected from thirty-six contemporary literary works in Chinese. The anaphora relations have been hand-annotated by linguists. The dataset is part of CLUE (Chinese Language Understanding Evaluation), a collection of Chinese language benchmarks analogous to GLUE \citep{xu2020clue}

The Indic General Language Understanding Evaluation (IndicGLUE) \citep{kakwani2020inlpsuite} includes translations of GLUE's WNLI into Hindi, Marathi, and Gujarathi.
Slovene translation of SuperGLUE benchmark includes the translation of \textsc{Wnli} into Slovene~\citep{zagar2020WSCslovene}.

Vered Shwartz has translated the WSC collection into Hebrew.\footnote{\url{https://vered1986.github.io/resources/winograd_he.html} \\
and 
\url{https://vered1986.github.io/resources/winograd_he.jsonl}} WSC has been translated into Russian \citep{shavrina2020russiansuperglue}. 
WSC273 and some other anaphora resolution datasets have been translated into Hungarian \citep{vadasz2022winograd}.

\subsection{Definite Pronoun Resolution Dataset}

The Definite Pronoun Resolution (\textsc{Dpr}) dataset is an easier variation of the Winograd Schema Challenge~\citep{Rahman2012DPR}.
The constraints on the Winograd schemas have been relaxed, and several examples in the dataset are not \textit{Google-proof}.
The dataset consists of $1322$ training examples and $564$ test examples, constructed manually.
$6$ examples in the training set reappear in \textsc{Wsc273} in a very similar form. 
These should be removed when training on \textsc{Dpr} and evaluating on \textsc{Wsc273}.
This dataset is also referred to as \textsc{WscR}, as named by~\citet{Opitz2018WSCRanking}.

An expanded version of this dataset, called \textsc{WinoCoref}, has been released by~\citet{Peng2015WinoCoref}, who 
further annotate all previously ignored mentions (in their work, a mention can be either a pronoun or an entity) in the sentences that were not annotated in the original work.
In this way, they add $746$ mentions to the dataset, $709$ of which are pronouns.



\subsection{Winograd Natural Language Inference Dataset}
\label{Wnli-section}
The Winograd Natural Language Inference (\textsc{Wnli}) dataset is part of the GLUE benchmark~\citep{Wang2019Glue} and is a textual entailment variation of the Winograd Schema Challenge.
An example from \textsc{Wnli} is given below with the goal to determine whether the hypothesis follows from the premise.

\smallskip 
\noindent\textbf{Premise: }\textit{The city councilmen refused the demon\-strators a permit because they feared violence.}

\noindent\textbf{Hypothesis: } \textit{The demonstrators feared violence.}

\noindent\textbf{Answer: } true / \textbf{false}

\smallskip 

The dataset consists of $634$ training examples, $70$ validation examples, and $145$ test examples.
Training and validation sets contain a major overlap with the \textsc{Wsc273} dataset, while test samples come from a previously unreleased collection of Winograd schemas.
Not all examples in this dataset contain the \textit{special} word and therefore do not come in pairs.
\citet{Kocijan2019MaskedWiki} note that examples are much easier to approach if the Winograd schemas are transformed from the textual entailment back into the pronoun resolution problem, and approached as such.

The same collection of examples is used for the SuperGLUE benchmark \citep{Wang2019SuperGLUE} as a pronoun resolution problem to begin with.
For the purpose of this survey paper, \textsc{Wnli} and SuperGlue \textsc{Wsc} are considered the same dataset.
They consist of the same examples and all approaches to \textsc{Wnli} described in this paper transform the examples as noted in the previous paragraph.

\subsection{WinoGender Dataset}

Unlike the previous 
datasets, \textsc{WinoGender} was created as a diagnostic dataset and is aimed to measure gender bias of the systems for pronoun resolution~\citep{Rudinger2018WinoGender}.
\textsc{WinoGender} consists of $120$ hand-written sentence templates, together with candidates and pronouns that can be inserted into the templates to create valid sentences.

In each sentence, one of the candidates is an occupation, usually one with a high imbalance in gender ratio (e.g., surgeon).
The other candidate is a participant (e.g., patient) or a neutral \textit{someone}.
For each sentence, either of the pronouns \textit{he, she}, or \textit{they} can be included to create a valid sentence, as the candidates are gender-neutral.
All together, this gives $720$ Winograd schemas.
An example from the dataset is 

\textit{The surgeon operated on the child with great care; [\textbf{his}/\textbf{her}] [tumor/affec\-tion] had grown over time.}

Note that the gender of the pronoun does not affect the expected answer; however, a biased system that associates the pronoun \textit{his} with the surgeon is likely to answer one of them incorrectly.
The aim of this dataset is not to measure model performance, as its data distribution is highly skewed, but to help analyse the models for gender bias.

A Swedish-language version of \textsc{WinoGender} has also been developed \citep{hansson2021swedish}.

\subsection{WinoBias Dataset}

\textsc{WinoBias} was created 
by~\citet{Zhao2018WinoBias}, which tries to identify gender bias in pronoun resolution models.
\textsc{WinoBias} and \textsc{WinoGender} were created concurrently but independently, despite the same objective.
They introduce a dataset with $3,160$ sentences, split equally into development and test.
Each sentence contains two candidates that are selected from a list of jobs with highly imbalanced gender ratio.

Two different templates are used to create Winograd schemas.
Type 1 sentences follow a structure that does not give away any syntactic cues. 
The authors thus estimate these sentences to be more challenging. 
An example of such a sentence is 

\textit{The farmer knows the editor because [\textbf{he}/\textbf{she}] [is really famous/likes the book].}

Type 2 sentences can be answered based on the structure of the sentence. 
The authors thus expect the models to perform better.
An example of such a sentence is 

\textit{The accountant met the janitor and wished [\textbf{her}/\textbf{him}] well.}

Just like in example of type 1, the referent is not affected by the grammatical gender of the pronoun.
Its ``twin pair'' has the candidates swapped.
As the structure of the sentence gives the answer away, there is no \textit{special~word}.

Moreover, the authors evenly split the whole dataset into \textit{pro-stereotypical} and \textit{anti-stereotypical}, depending on whether the gender of the pronoun matches the most common gender of the referent occupation or not.
They observe that publicly available models for co-reference resolution exhibit a major difference (up to $21.1\%$ $F_1$) in performance on \textit{pro-} and \textit{anti-} subsets of the dataset.

\subsection{WinoGrande Dataset}
\label{Section:Winogrande}
In contrast to the sets of Winograd schemas discussed above, created either entirely by hand or using templates and a fixed set of rules for instantiating the templates, and consisting of anywhere from a few hundred to a few thousand examples, \citet{Sakaguchi2020WinoGrande} presented a substantially larger ($\sim$44K) set of coreference resolution problem sentences, along with methods to filter these sentences to remove bias.  As discussed in~\citep{ morgenstern21importanceofwinogrande},  Sakaguchi et al.\ were primarily motivated by the fact that systems already existed that achieved near-human performance on the Winograd Schema challenge, yet none had achieved the ability to do commonsense reasoning, a central aim of posing the WSC. They therefore sought to re-engineer the WSC as a more meaningful benchmark of a system’s CSR ability, by introducing new methods of dataset development and adversarial filtering, expressly designed to prevent AI systems from making claims of smashing through benchmarks without making real progress.

The  \textsc{WinoGrande} corpus was collected using the Amazon Mechanical Turk (MTurk) crowd-sourcing marketplace. To prevent crowd workers from creating lexically and stylistically repetitive examples, they were primed by a randomly chosen topic from a WikiHow article as a suggestive context. Additionally, they were instructed to adhere to constraints close to the original WSC constraints, e.g., avoiding writing examples where word association would lead to inferring the correct pronoun referent. Workers were told to focus on the domains of social common sense and physical common sense. An additional set of Mechanical Turkers checked all sentences generated to increase the likelihood that humans could easily infer pronoun referents but that making such inferences required at least some commonsense knowledge.


 
The authors additionally introduced the \textsc{AfLite} adversarial filtering algorithm.The idea was to retain only examples that minimize representation bias. Removed pairs include those with data-set specific polarity basis (e.g., advanced rock climbing is more strongly associated with being {\em strong} than being {\em weak}). 
They used a fine-tuned RoBERTa language model~\citep{Liu2019Roberta} to gain contextualized embeddings for each instance.
Using these embeddings, they iteratively trained an ensemble of linear classifiers, trained on random subsets of the data and discard top-$k$ instances that were correctly resolved by more than $75\%$ of the classifiers.
By iteratively applying this algorithm, the authors identified a subset ($12,282$ instances), called 
$\mbox{WinoGrande}_{\mbox{debiased}}$.
Finally, they split this dataset into training ($9,248$), development ($1,267$), and test ($1,767$) sets. In many cases, the AFLITe filtering removes only one sentence of a pair. Slightly more than 1/2 of the questions in $\mbox{WinoGrande}_{\mbox{debiased}}$ are not twins.\footnote{\citet{Sakaguchi2021WinoGrande} state that 1/3 are not twins, but our own count found that the true figure
is 56\%.}
They also released the unfiltered training set 
$\mbox{WinoGrande}_{\mbox{all}}$
with $40,938$ examples \citep{Sakaguchi2020WinoGrande}; these are all twins.

Despite the promise of this approach, we note first, that many of the sentences in the filtered \textsc{WinoGrande} corpus do not fit the criteria laid out in the WSC, and therefore are not actually Winograd schemas. Many sentences are easily solvable using word  correlation; that is, they are not Google-proof. For example (all examples below are from the debiased, filtered set released in September 2020), consider: {\em The doctor diagnosed Justin with bipolar and Robert with anxiety. Justin/Robert had terrible nerves recently.} A Google search shows that {\em anxiety} and {\em nerves} are more strongly associated than {\em bipolar} and {\em nerves}. Some sentences appear to be poorly written or to contain the answer directly in the sentence, e.g. {\em The waiter could not cover the round tables with the square tablecloths because the tables/tablecloths were square.} Others are genuinely hard to understand, e.g., {\em George opted for both of them to use a knife instead of a gun in the duel because the knife/gun could partially injure them.}

We note, second, that the goal of postponing human-level performance on this dataset until systems achieve commonsense reasoning ability was achieved only briefly. The \textsc{WinoGrande} leaderboard shows that the UNICORN system \citep{DBLP:conf/aaai/LourieBBC21} achieves 91.2\% accuracy, despite little progress in commonsense reasoning.

\subsection{WinoFlexi Dataset}
Similarly to \textsc{WinoGrande},~\citet{Isaak2019WinoFlexi} aimed to construct a dataset through crowdsourcing.
They built their own crowdsourcing interface for generating and evaluating schemas, and they collected $165$ candidate Winograd schemas. In our manual review of this, we found that there were 84 distinct valid schemas ($168$ examples).
Unlike workers on \textsc{WinoGrande}, workers on \textsc{WinoFlexi} were not presented with any particular topic and were free to pick it on their own; and they were not instructed to avoid examples that could be solved using word associations.  

\subsection{Winventor Dataset}
The Winventor program~\citep{isaak2020winventor} uses automated methods
to generate candidates for pronoun disambiguation problems, which could then be polished or adapted by human post-editors. Using hand-coded techniques, the program collected Wikipedia sentences with a pronoun with two potential referents and formulated a question based on the resolution of that pronoun. (An alternative, less successful, program tried to use a deep learning architecture for the same task.) 

For example, starting with the Wikipedia sentence, ``As Frederick was rather distant to his family, Eleanor had a great influence on the raising and education of Frederick's children, and she therefore played an important role in the House of Hapsburg's rise to prominence,'' Winventor proposed the question ``Who therefore played an important role in the House of Hapsburg's rise to prominence?'' Since, as it stands, ``she'' can be disambiguated by gender constraints, a human editor changed ``Eleanor'' to ``John'' and ``she'' to ``he''. The authors compared the performance of human experts in generating PDPs when aided by Winventor as compared to working by themselves; they found that, using Winventor the editors were able to generate examples 1.5 times as fast as without it (10 PDPs in 20 minutes as opposed to 7), and that their examples were much more varied in form.

\subsection{Datasets of Explanations}
\textsc{WinoWhy}~\citep{zhang2020winowhy} and \textsc{WinoLogic}~\citep{he2021winologic} are collections of correct and incorrect natural language explanations of all Winograd Schemas in the \textsc{Wsc273} dataset.
Moreover, \citet{Yordan2021eWinogrande} collect a small set of explanations for the \textsc{WinoGrande} dataset, called \textsc{e-WinoGrande}.
The goal of all three datasets is to determine whether systems that can correctly answer Winograd Schemas are also capable to identify the correct explanations for their choice.
The answer to this question at the time of the writing seems to be negative.
The difference between the datasets is in their construction. WinoWhy and \textsc{e-WinoGrande} were constructed through crowdsourcing with no guarantee that the explanations are exhaustive, that is, a correct explanation is not guaranteed to contain all the necessary steps to get the answer.
WinoLogic, on the other hand, consists of explanations that were obtained by manually deriving the answers to examples in first-order logic first and then rewriting the derivation into natural language explanations.

\subsection{Winograd Schemas for Machine Translation}
Machine translation of schemas to languages with a strong presence of grammatical gender can require their implicit resolution. For instance,
\citet{emelin2021winox} provide an example of the following translation from English to German:
\begin{quote}
    The gardener used the shovel more than the rake because \textit{it} was poorly made.
    
    Der G\"{a}rtner benutzte die Schaufel mehr als den Rechen, weil [\textit{er}/sie] schleht gemacht war.
\end{quote}
\textit{Die Schaufel} and \textit{der Rechen} have feminine and masculine grammatical gender, respectively, and to correctly translate the pronoun \textit{it} into \textit{sie} or \textit{er}, the system must resolve the pronoun first.
Alternatively, \citet{stanovsky2019BiasMT} note that translating the sentence
\begin{quote}
    The doctor asked the nurse to help \textit{her} in the procedure.
\end{quote}
requires implicit resolution of the pronoun to determine whether to use the feminine or masculine form of \textit{the doctor} (in German, \textit{der Arzt/die \"{A}rztin}).

\citet{stanovsky2019BiasMT} create a \textsc{WinoMT} dataset to detect gender bias in machine translation systems by observing how sentences in WinoGender and WinoBias are translated to Spanish, French, Italian, Russian, Ukrainian, Hebrew, Arabic, and German.
\citet{kocmi2020winomt} extend this to Polish and Czech.
They find that systems for machine translation regularly assign grammatical gender based on gender stereotypes.
\citet{emelin2021winox}, on the other hand, focus on the subset of WinoGrande that talks about inanimate objects that can be referred to with the pronoun \textit{it}.
They create a Wino-X, a diagnostic dataset for machine translation from English to French, German, and Russian. They, too, find that existing machine translation systems do not resolve Winograd Schemas well.
Curiously, they find that systems are more likely to (incorrectly) translate an entity into a masculine than feminine form, indicating that gender bias and possible imbalance in the training data affects even the treatment of inanimate objects with feminine grammatical gender.

\subsection{Winoground Dataset}
\begin{figure}[h]
    \centering
    \includegraphics[width=4in]{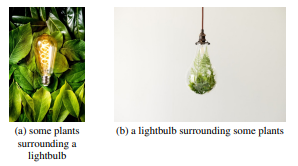}
    \caption{Example from Winoground \citep{thrush2022winoground}}
    \label{figWinoground}
\end{figure}

Winoground \citep{thrush2022winoground} is a challenge dataset for visio-linguistic compositional reasoning. Each example consists of a pair of images and a pair of captions that differ only in word order; the task is to match the correct caption with the correct image (Figure~\ref{figWinoground}).

\subsection{An earlier dataset for psycholinguistic experimentation}
In psycholinguistic studies of pronoun disambiguation, \cite{kehler2008coherence} report using experimental material that includes a set of 16 pairs of Winograd schemas following a fixed framework:
\begin{quote}
Samuel threatened Justin with a knife, and Erin [blindfolded/stopped] him.
Whom did Erin [blindfold/stop]?

Samuel threatened Justin with a knife, and he [blindfolded Erin/alerted security].
Who [blindfolded Erin/alerted security]?
\end{quote}

\subsection{Evaluation with Human Subjects}
Human subject evaluation has been carried out for a number of data sets (see Table~\ref{tabHumanSubjects}).
The study reported in \citep{kehler2008coherence} were carried out by undergraduates at UCSD, who were self-reported monolingual English speakers.
The studies reported in \citep{davis2016human} and the unpublished studies carried out by Davis were carried out in person by paid volunteers from the NYU community (in the latter case, the volunteers were screened for being native speakers of English).
In all the other studies, the subjects were crowd workers recruited on Amazon Mechanical Turk or similar platforms. The details of the methodology used are given in the references. The anomalously low
result of 86.5\% accuracy on WSC273 is mentioned in passing in a footnote in \citep{Rudinger2018WinoGender}; the method used is not described.

\begin{table}
    \centering
    \begin{tabular}{l|l|l}
{\bf Dataset}& {\bf \% correct} & 
{\bf Reference} \\ \hline
Experimental dataset of & 93.0\% &
\citep{kehler2008coherence} \\
32 schemas & & \\ \hline
143 schemas from WSC273  & 92.1\%
& \citep{bender2015establishing} \\ \hline
66 texts with 108 PDPs & 90.1\% &
\citep{davis2016human} \\ \hline
89 unpublished Winograd schemas &  92\% & \citep{davis2016human} \\ \hline
Unpublished collection of & 96\% & Unpublished test, \\
86 texts with 101 PDPs  & & (Ernest Davis) \\
 \hline
WNLI &  96.1\% & \citep{nangia2019human}
\\ \hline
{\sc WinoGender} & 94.9\% & \citep{Rudinger2018WinoGender} \\ \hline
{\sc WSC273}  & 86.5\% & \citep{Rudinger2018WinoGender} \\ \hline
{\sc WinoGrande}  & 94\% & \citep{Sakaguchi2020WinoGrande} \\ \hline
    \end{tabular}     
    \caption{Human subject evaluation of datasets}
    \label{tabHumanSubjects}
\end{table}

\section{Overview of Methods to Solving the Winograd Schema Challenge}
\label{appendix-methods}
\subsection{History of pronoun disambiguation in AI and linguistics}

The AI problem of building natural language understanding systems capable of determining the referent of pronouns in text,\footnote{As with many linguistic phenomena, the issue is much less crucial in generation than in interpretation, because a generator generally can avoid difficult cases by limiting its pronoun use to clear-cut cases, with at worst some loss in the naturalness of the expression.} and the related linguistics problem of characterizing the rules that govern pronoun reference have been studied extensively since the 1970s.

Pronouns in English are generally, though not always, anaphoric---that is, a pronoun generally refers to some entity mentioned by a noun phrase earlier in the text---and most of the AI research on pronoun use has focused on finding anaphoric references in the text, since those are more easily characterized than non-anaphoric references. The problem is thus a subset of the broader problem of anaphoric resolution. Indeed, in general, pronoun resolution is a particularly easy category of anaphoric resolution for computational systems; the problem of determining whether two noun phrases in text refer to the same entity is often much more difficult (Claire Cardie, personal communication). (Extensive surveys of anaphoric reference are given in \citep{poesio2011computational},  \citep{poesio2016anaphora}, and \citep{kehler2008coherence}.)

Early AI work in pronoun resolution mostly focused on the role of commonsense knowledge in disambiguation (e.g.,  \citep{charniak1972toward, wilks1975intelligent, schank1977scripts, hobbs1979coherence, hobbs1993interpretation}) or on structural characteristics of the text, such as parallelism and focus \citep{grosz1977representation, sidner1979towards, kameyama1986property}. Implementations were almost always proof-of-concept rather than practical systems. Development was guided by the researchers' intuitions and interests, rather than by evaluation. Systematic evaluation of systems was extremely rare \citep{mcdermott1976artificial}. 

As we have discussed in Section~\ref{secConstructingWSs}, it is clear that both types of information are used in human interpretation of natural language. When different constraints conflict, the rules for resolving the conflict are complex: sometimes pragmatic preferences overrule formal preferences, sometimes {\em vice versa,} sometimes the conflict leads to an unintelligible sentence. This line of work continues among linguists \citep{kehler2008coherence}, but has largely been abandoned in AI, due to the difficulties of developing such systems at scale.

\medskip 
\noindent {\bf Statistical approaches:} The early DARPA-funded conferences MUC-6 and MUC-7 (Message Understanding Conferences) \citep{grishman1996design, chinchor1998overview} included an evaluation of coreference resolution systems. Both conferences involved the construction of large text corpora, drawn from natural sources, and carefully hand-annotated following guidelines that were carefully worked out (though not uncontroversial) and presented in careful detail to the annotators. Systems used the machine learning technology available at the time and were both trained and evaluated on these corpora. Systems were generally trained for a fairly narrowly defined tasks; progress relative to the evaluation metric tended to take place steadily, in small increments. This general approach dominated work in the area for the next~decade.

\medskip 
\noindent {\bf Deep learning systems trained on big data:} A second revolution in natural language processing began in the early 2010's, and has dominated the field since around 2016. The technology used is some variant of deep learning; in the last two years, particularly transformer-based technology \citep{Devlin2019Bert}. Much of the most impactful technology is task-independent, particularly enormous language modeling systems and word embeddings; performance on individual tasks can then be enhanced with fine-tuning. Data sets of all kinds---enormous unannotated corpora, datasets annotated or created by crowd-workers, synthetic datasets---have proliferated. Progress has in some cases been astonishing.

In statistical approaches, and still more in approaches based on deep learning, commonsense knowledge has been incorporated into the systems only to the extent that the learning procedure implicitly abstracts it from text corpora.

Though the deep-learning technology had led to incomparably more powerful systems than the knowledge-based approach, the former does not subsume the latter. Many of the issues that were addressed in knowledge-based systems are not reliably solved in 2021 state-of-the-art technology. For instance, the language modeling system GPT-3 failed \citep{Marcus2020GPT3} when tested on an example from \citep{charniak1972toward}.

\subsection{Feature-based Approaches to the Winograd
Schema Challenge}

This section covers the approaches that collect knowledge in form of explicit rules from knowledge bases, internet search queries, and use logic-based systems or optimization techniques to deduce the answer.
Results of methods that rely on search engines, such as Google, can be irreproducible, 
as they strongly depend on the search~results.

The majority of the early approaches to the Winograd Schema Challenge were feature-based or logic-based.
While powerful in theory, their main bottleneck was usually the processing of the input data or relevant background knowledge.
This is best demonstrated by \citet{Sharma2019ASP3} who designs a reasoning algorithm that solves $84.2\%$ of examples in $WSC285$ if the input processing and background knowledge acquisition are done manually.
The same algorithm solves fewer than $50\%$ of examples when data processing is done with K-Parser~\citep{Sharma2015Kparser}.

At the same time, the methods in this group are the hardest to compare one to another.
The models are often evaluated on specific subsets of \textsc{Wsc273} or other datasets, making direct comparison hard.
We thus chose not to summarize the results in a table like we do for neural and language-model-based approaches.

Approaches in this work usually work in three steps: First, the input sentence is processed to extract keywords and semantics. In the second step, relevant background knowledge is extracted from various sources, most commonly knowledge bases and internet search queries. Finally, all this information is combined used the input to the \textit{reasoning algorithm},  which gives the final answer.
The third step is where approaches differ the most one from another.
Approaches used were, for example, SVM-rankers~\citep{Rahman2012DPR}, integer linear programming~\citep{Peng2015WinoCoref}, answer set programming~\citep{Sharma2015Kparser}, message passing on a graph~\citep{Fahndrich2018MarkerPassing}, and formal logic~\citep{Isaak2016WinoSense}.

While many of these models achieve decent results on the data they have been evaluated on, their ability to generalize can be contested.
\citet{Sharma2015Kparser}, for example, report achieving $69\%$ on a selected subset of \textsc{Wsc285}, but when re-evaluated on a different subset of the data, the same model only achieved chance-level performance~\citep{Zhang2018DistributedWSC}.

We highlight \citep{Emami2018KnowledgeHunter} as the first model to achieve a better-than-chance accuracy ($57.1\%$) performance on the entire \textsc{Wsc273}.
Their system is completely rule-based and focuses on high-quality knowledge hunting, rather than reasoning, showing the importance of the former.
Unlike neural approaches from later sections, this model is not negatively affected by switching candidates.



With the increase in the popularity of large language models, interest in feature and reasoning-based methods decreased, however, \citet{hong2020tackling} note that hybrid approaches can be useful when solving schemas of a very specific type, using a language model when the symbolic approach does not yield an answer.

\subsection{Neural Approaches}
\label{section-Neural}
This section contains approaches that rely on neural networks and deep learning, but do not use pre-trained language models.
Models in this section are usually designed, built, and trained from scratch in contrast to models that use language models and are usually built on top of an off-the-shelf pre-trained neural network.
We find that several ideas introduced in this section are later adjusted and scaled to language models; see~\ref{section-LM}.
Note that each work comes with a collection of mo\-del-specific architecture designs that are not covered in~detail.

\begin{table}[]
    \centering
    \begin{tabular}{@{}l@{\ }c@{\ }|@{\ }cc@{}}
         & training resource  & \textsc{Wsc273} & \textsc{Pdp}\\ \hline
        \cite{Liu2017NAM} & cause-effect pairs & $70\%\dagger$& -- \\
        \cite{Liu2017KEE} & cause-effect pairs, Ontonotes & $52.8\%$ & $58.3\%$ \\
        \cite{Zhang2018DistributedWSC} & Wikipedia, dependency parser & $60.33\%\dagger$ & -- \\
        \cite{Opitz2018WSCRanking} & DPR, InferSent & $56\%$ & --\\
        \citet{Wang2019UDSSM} & Gutenberg, 1 Billion Word & $62.4\%$&$78.3\%$ \\
    \end{tabular}
    \caption{Resources, both data and tools, used by different neural approaches, ordered by the time of publication.\\ $\dagger$ denotes results that were obtained on modified version of the \textsc{Wsc273} dataset, usually a hand-picked subset. Such results are not guaranteed to generalize to full dataset.}
    \label{NN-table}
\end{table}

We summarize the resources and approaches of neural methods in Table~\ref{NN-table} and give main observations in the following paragraph.
Most of the methods in this section attempt to incorporate \textit{semantic features} into the unsupervised training of the models.
The listed pieces of work usually introduce tricks that aim to force a model to pick up the relevant training signals, either through specific architectures, or through appropriate formulation of the training data.
Success is often limited and many models are thus only evaluated on subsets of the dataset that suit the model better, making a direct comparison of the results non-trivial.
An evident trend through time is the movement from methods based on \textit{(contextual) word embeddings} (e.g.,~\citep{Zhang2018DistributedWSC}) into end-to-end systems with a more complex set of objectives~\citep{Wang2019UDSSM}. 

\subsection{Language Model Approaches}
\label{section-LM}
This section covers the approaches that use neural language models to tackle the Winograd Schema Challenge.
Most of them use one or more language models that were trained on a large corpus of text.
Several authors use large pre-trained language models, such as BERT~\citep{Devlin2019Bert}, and have to tailor their approach accordingly.
Such works thus focus on better fine-tuning of such language models instead of inventing new architectures.

\begin{table}[]
    \centering
    \begin{tabular}{@{}l|c@{}c@{}}
         & Language Model & fine-tuning or external data \\ \hline
        \citep{Trinh2018WinogradGoogle} & custom LSTM & -- \\
        \citep{Radford2019GPT2} & GPT-2 & -- \\
        \citep{Klein2019AttentionWSC} & BERT & -- \\
        \citep{Prakash2019KnowledgeHuntingLMs} & custom LSTM & internet querying \\
        \citep{Kocijan2019MaskedWiki} & BERT & MaskedWiki, DPR \\
        \citep{Kocijan2019WikiCREM} & BERT & WikiCREM, DPR, GAP \\
        \citep{Ruan2019WSCNSP} & BERT & DPR \\
        \citep{He2019HNN} & BERT & DPR \\
        \citep{Ye2019AMS} & BERT & ConceptNet, DPR \\
        \citep{Sakaguchi2020WinoGrande} & RoBERTa & WinoGrande \\
        \citep{Melo2020WSCportugese} & custom LSTM & -- \\
        \citep{Brown2020GPT3} & GPT-3 & -- \\
        \citep{yang2020generative} & RoBERTa & WinoGrande and generated data \\
        \citep{lin2020tttttackling} & T5 (3B) & WinoGrande \\
        \citep{khashabi2020unifiedqa} & T5 & WinoGrande and QA tasks \\
        \citep{Lourie2021Unicorn} & T5 & WinoGrande and RAINBOW \\
    \end{tabular}
    \caption{Resources used by different language-model based approaches, ordered by the time of publication. With time, ever larger language models and more additional fine-tuning data was used.}
    \label{LM-method-table}
\end{table}

\begin{table}[]
    \centering
    \begin{tabular}{@{}l|cccc@{}}
         & \textsc{Wsc273} & \textsc{Wnli} & \textsc{Pdp} & \textsc{WinoGrande} \\ \hline
        \citep{Trinh2018WinogradGoogle} & $63.7\%$ & -- & $70\%$ & -- \\
        \citep{Radford2019GPT2} & $70.7\%$ & -- & -- & -- \\
        \citep{Klein2019AttentionWSC} & $60.3\%$ & -- & -- & -- \\
        \citep{Prakash2019KnowledgeHuntingLMs} & $70.17\%$ & -- & -- & -- \\
        \citep{Kocijan2019MaskedWiki} & $72.5\%$ & $74.7\%$ & -- & -- \\
        \citep{Kocijan2019WikiCREM} & $71.8\%$ & $74.7\%$ & $86.7\%$ & -- \\
        \citep{Ruan2019WSCNSP} & $71.1\%$ & -- & -- & -- \\
        \citep{He2019HNN} & $75.1\%$ & $89\%$ & $90.0\%$ & --\\
        \citep{Ye2019AMS} & $75.5\%$ & $83.6\%$ & -- & -- \\
        \citep{Sakaguchi2020WinoGrande} & $90.1\%$ & $85.6\%$ & $87.5\%$ & $79.1\%$ \\
        \citep{Brown2020GPT3} & $88.3\%$ & -- & -- & $77.7\%$ \\
        \citep{yang2020generative} & -- & -- & -- & $80.0\%$ \\
        \citep{lin2020tttttackling} & -- & -- & -- & $84.6\%$\\
        \citep{khashabi2020unifiedqa} & --& -- & -- & $89.4\%$ \\
        \citep{Lourie2021Unicorn} & -- & -- & -- & $91.3\%$ \\
    \end{tabular}
    \caption{Results of language-model based approaches on the most commonly used datasets. Less commonly used evaluation sets include \textsc{Dpr} and \textsc{Wsc285}. The only non-English result by~\citet{Melo2020WSCportugese} who evaluate their model on a set of Portugese Winograd Schemas is not included.}
    \label{LM-results-table}
\end{table}

We summarize the resources and results of all language-model based approaches in Tables \ref{LM-method-table} and \ref{LM-results-table}, respectively.
A careful comparison of the approaches in the listed works reveals multiple changes of trends over time.
We list and discuss them below.
\begin{itemize}
    \item Used language models consistently rise in complexity. First approaches use custom-made LSTMs, trained specifically for the task, which are replaced with large-scale language models such as BERT, in turn replaced with even larger T5 and GPT-3. This trend is not exclusive to the Winograd Schema Challenge, and it is a general ongoing trend in~NLP. 
    \item Appropriate fine-tuning grows in importance, as it turns out to be the most efficient way of improving performance. 
    Approaches that use language models without fine-tuning (e.g.,  \citep{Trinh2018WinogradGoogle} and \citep{Klein2019AttentionWSC}) are outperformed by approaches that fine-tune the same models on WSC-like data (e.g., DPR). To increase the impact of fine-tuning, language models are often fine-tuned on more than one dataset. Earlier approaches usually use synthetic datasets~\citep{Kocijan2019MaskedWiki,Ye2019AMS,yang2020generative}, while later approaches resort to multi-task learning on related tasks~\citep{khashabi2020unifiedqa,Lourie2021Unicorn}.
    \item \textsc{Wnli} only becomes commonly used after the reformulation trick described in \ref{Wnli-section} is introduced by~\citet{Kocijan2019MaskedWiki}. We note that many models were evaluated on this dataset as part of the \textsc{Glue} benchmark~\citep{Wang2019Glue}, which we do not include in this survey,  because they do not introduce any ideas specific to the task. Their novelty is in better pre-training, which is not the focus of this paper.
    \item The introduction of WinoGrande dataset made most other smaller datasets less interesting to researchers, even if these small datasets are higher in quality. The majority of the papers published after the release of WinoGrande~\citep{Sakaguchi2020WinoGrande} only evaluated their work on that dataset.
\end{itemize}

We highlight that many of the approaches that use large-scale pre-trained language models use various input and output formats, which can affect the performance of the model.
For a detailed comparison of different training objectives and their impact on the performance of the pre-trained language model, we refer the reader to the work by~\citet{Yordanov2020Objective} and~\citet{liu2020formalization}.

\section*{Acknowledgments}
This work was supported by the Alan Turing Institute under the EPSRC
grant EP/N510129/1, the AXA Research Fund, the EPSRC
grant “Unlocking the Potential of AI for English Law”, and
the EPSRC Studentship OUCS/EPSRC-NPIF/VK/1123106.

%
%
%

\bibliographystyle{elsarticle-num-names-alpha}

\bibliography{references.bib}
\end{document}